\documentclass[letterpaper, 10 pt, conference]{ieeeconf}
\IEEEoverridecommandlockouts                              % This command is only needed if 
\usepackage{cite}
\usepackage{amsmath,amssymb,amsfonts}
\usepackage{algorithmic}
\usepackage{graphicx}
\usepackage{textcomp}
\usepackage{xcolor}
\usepackage{booktabs}
\usepackage{subcaption}
\usepackage{placeins}
\usepackage{float}
\usepackage{multirow}
\usepackage{multicol}
\let\labelindent\relax
\usepackage{enumitem}

\def\BibTeX{{\rm B\kern-.05em{\sc i\kern-.025em b}\kern-.08em
    T\kern-.1667em\lower.7ex\hbox{E}\kern-.125emX}}

\usepackage[pagebackref,breaklinks,colorlinks]{hyperref}

\title{\LARGE \bf
Synthesizing and Identifying Noise Levels in Autonomous Vehicle Camera-Radar Datasets
}

\author{Mathis Morales$^{1}$ and Golnaz Habibi$^{2}$% <-this % stops a space
\thanks{University of Oklahoma}% <-this % stops a space
\thanks{$^{1}$Mathis Morales, School of Electrical and Computer Engineering,
        University of Oklahoma, Norman, Ok, USA
        {\tt\small mathis.morales@ou.edu}}%
\thanks{$^{2}$Golnaz Habibi, School of Computer Science,
        University of Oklahoma, Norman, Ok, USA
        {\tt\small golnaz@ou.edu}}%
}

\begin{document}

\maketitle

\begin{abstract}
Detecting and tracking objects is a crucial component of any autonomous navigation method. For the past decades, object detection has yielded promising results using neural networks on various datasets. While many methods focus on performance metrics, few projects focus on improving the robustness of these detection and tracking pipelines, notably to sensor failures. In this paper we attempt to address this issue by creating a realistic synthetic data augmentation pipeline for camera-radar Autonomous Vehicle (AV) datasets. Our goal is to accurately simulate sensor failures and data deterioration due to real-world interferences. We also present our results of a baseline lightweight Noise Recognition neural network trained and tested on our augmented dataset, reaching an overall recognition accuracy of 54.4\% on 11 categories across 10086 images and 2145 radar point-clouds. The code for this project is available \href{https://github.com/airou-lab/SynthCRAV}{[here]}.
\end{abstract}

\section{Introduction}
To achieve autonomous navigation, vehicles need a way to perceive their surrounding environments. For a ground vehicle, for instance, driving in a complex environment with other cars requires it to dynamically observe its surroundings to avoid colliding with another car or a pedestrian for example, or simply to respect the law and stop at a stop sign. The most common sensors used in autonomous vehicles are Cameras, LiDARs, Radars, ultrasonic sensors and infrared sensors \cite{review2024}.\\
Cameras are essential as they provide semantic information unmatched by other sensors. For that reason, a lot of research focuses on Camera-based object detection. Current State of The Art (SOTA) methods achieve high object recognition accuracy on 2D detection tasks, such as co-DETR \cite{detr} reaching 66 \% accuracy on the COCO dataset \cite{coco} for 80 object categories accumulated on 330k images.\\
However, as ground-based and air-based vehicles move in a three dimensional environment, and are surrounded by 3D objects, they need to be able to accurately detect the position of objects in 3 dimensions, as well as their categories, orientations, dimensions, and velocities. All of these are important to be able to correctly estimate the motion profile of a detected object to avoid it or to track it. Unfortunately, for all the semantic information they provide, cameras can only output a two dimensional pixel map. This makes it extremely challenging to correctly estimate the depths, distances, and shapes of objects using only this sensor. Furthermore, The challenges of 3D detection and tracking create a need for three dimensional datasets. In autonomous navigation, the nuScenes dataset \cite{caesar2020nuscenes} is a reference, as it provides users with a Camera-Radar-LiDAR dataset, focused on urban environments, with benchmarks for 3D Detection tasks across 10 categories and for 3D tracking on 7 of them. It is also considered to be one of the first large-scale dataset of the sort, with 1000 scenes of 20 seconds, gathering over 1.4 million images \cite{review2024}.\\

As mentioned earlier, Camera-only detection networks have more difficulties to accurately place 3D bounding boxes. This is highlighted in the nuScenes 
leaderboard, with the highest-ranking camera-only method \cite{sparse4d} reaching 75th place on the Detection task out of 338.\\
To increase detection performances, a depth-measuring sensor is required, such as Radars or LiDARs. Most SOTA methods prefer using LiDAR systems for their precision \cite{ab3dmot, caesar2020nuscenes, centerpoint}. In fact, if we consider the nuScenes leaderboard the first 67 detection methods use the car's LiDAR, performing better than the SOTA Camera-Radar method \cite{RCBEVDet++}. On the tracking task, the first Camera-Radar method \cite{RCBEVDet++} ranks 75th out of 128.\\
However, LiDARs are expensive, heavy, large, hard to use, memory-demanding, and difficult to integrate in existing vehicles. Real-time processing would also require large computational power, leading to significant battery consumption. For this reason, other methods prefer to focus on RADAR technology instead. Radars have been used for almost a century in a variety of applications \cite{lang2020comprehensivesurveymachinelearning, principlesofmodernradar, reviewautomotiveradars}, making them a well-known and cheap sensor, especially compared to LiDARs. They also offer unequaled range measurement accuracy, velocity measurement accuracy, and robustness to adverse weather conditions like fog, rain or snow, in which the LiDAR systems perform poorly \cite{dreissig2023survey}.
The downside is Radars produce a very sparse data point-cloud, which can make Radar detection challenging. This is, however, mitigated by the Camera's properties \cite{Long_Kumar_Morris_radiant, kim2023crn}.\\

For real-life Autonomous Vehicles, robustness must be a priority, as we cannot afford unreliable measurements while a car is driving on the street and people's lives are at stake. For this reason we focus on Camera-Radar sensor fusion and consider the following argument:\\
As models are getting better at performing Detection and Tracking on offline datasets that have been curated, cleaned, and annotated, how would they fare if deployed in a real-world car? Furthermore, how would they react to a sensor that fails and sends erroneous data for various reasons?\\
The base assumption is a detection network may be able to deal with a certain amount of erroneous data, but would ultimately fail past a certain point. This argument is valid for any neural network method, as they all have a point of failure. Considering the scenario of a car driving next to a power transformer in the street. As the power transformer generates large Electro-Magnetic (EM) interferences, the radar sensor fails and begins sending wrong data to the autonomous vehicle. As this happens a car in front of it brakes, but the radar sensor estimates its position to be much further away than it actually is, and the autonomous vehicle collides with it. Another scenario is a car coming out of a bridge, the camera takes a little time to adjust its exposure level, during which the front camera is blinded by the light, potentially leading to a collision due to a missed detection\\
As these situations are possible, there must be safeguards set in place to increase robustness in case of degraded sensor conditions. This project aims at addressing how we can accurately estimate a sensor-failure degradation (or noise) level, which, to the best of our knowledge, has never been done before for autonomous vehicle on 3D camera-radar datasets.\\

The main contributions of this project are:
\begin{itemize}
    \item A data augmentation pipeline focused on synthesizing real-world sensor defects for a camera-radar autonomous vehicle dataset. 
    \item A baseline for noise recognition, using lightweight models to directly estimate the degradation level. This allows a plug-and-play use of our method, as any object detection can use the noise level information to take action.
\end{itemize}

\section{Related works}
As mentioned in \cite{review2024}, while most method focus on performance metrics, few focus on sensor robustness.\\
RadSegNet \cite{RadSegNet} independently extract information from each sensor, but only focuses on camera failure, showing their method can still work reliably using only the radar data. While the approach is similar to ours, our methods takes place before any feature extraction, allowing an easy switch to another detection method depending on the noise level. They also focus more on adverse weather conditions, which is different from sensor degradation that can be induced by EM interferences, thermal noise, or highly reflective surfaces (for radars).\\
On the other hand, methods such as ImmFusion \cite{ImmFusion} use Transformer networks to actively select useful information from the sensors. While generative models and self-attention mechanisms provide an interesting avenue to detect and deal with sensor defects, they still have a black-box effect, making it hard to guarantee robustness. They could also benefit from accurate representation of data degradation in training, which our method could help provide.\\
Furthermore, both methods couple objects detection with robustness improvements, leading to incompatibilities with existing detection methods.

\section{Methodology}
Dealing with both Camera and Radar data, we divide our methodology in two parts, one for each sensor. We create two noise level dials, both referred to as $N_{lvl}$. The two dials are handled separately and have different meaning for each sensor, but both control a noise level going from 0\% to 100\%. The noise levels can also be increased above these values for an even higher degradation simulation, but they have been tailored so that the 100\% noise level represent the highest realistic distortion in real life scenarios.

\subsection{Synthetic Camera Degradation}
We propose four different common degradation types for images: Blurring, Low exposure, High exposure, and Additive Noise.
\begin{enumerate}[label=\textbf{\alph*)}]
    \item \textbf{Blurring.} This happens when the camera lens improperly focuses lights from the scene onto the image plane. It's mostly due to a calibration issue, but can be induced by an auto-focus failure because of low light condition or misalignment due to a shock. Motion-blur is also similar to focus blur, and could be simulated by it to a certain extent.

    Blurring effect is generally simulated by a convolution with a Gaussian kernel. We synthesize different noise levels by changing the size of the Gaussian kernel:
    \begin{equation}
    k_{size}= 2 \times round(N_{lvl})+1
    \end{equation}
    Figure \ref{blur} shows the blur effect at different distortion levels.
    
    \begin{figure}[htbp]
    \centerline{\includegraphics[width=0.5\textwidth]{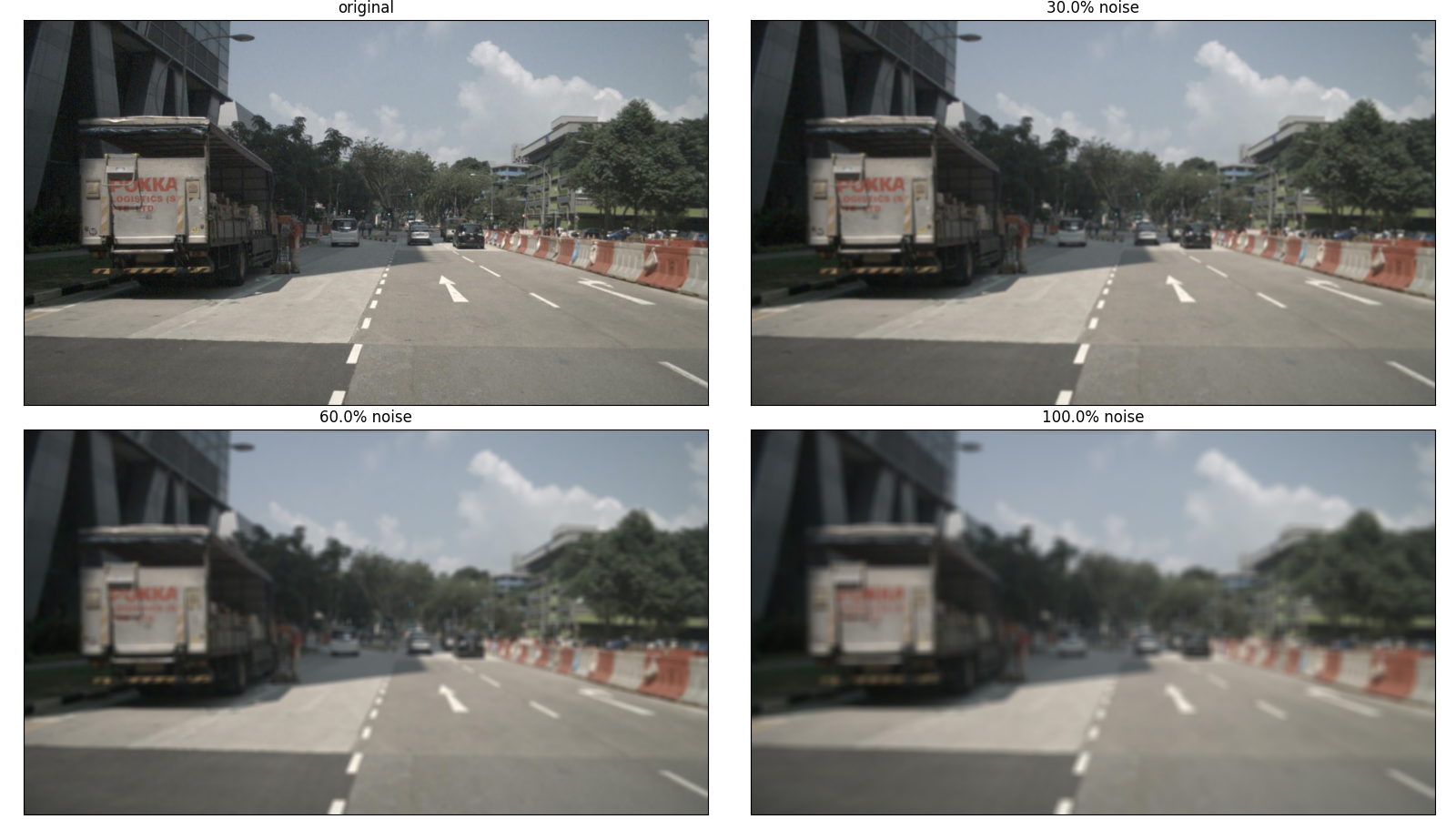}}
    \caption{Blur effect at 0\%, 30\%, 60\%, and 100\% (max) distortion levels.}
    \label{blur}
    \end{figure}
    
    \item \textbf{High/Low exposure.} Here we simulate the effect of a camera going in (low) or out of (high) a tunnel or shaded area. At this moment the light level changes abruptly, and the camera automatically adjusts the exposure level of the lens to capture more or less light. However, for a small instant of one or two frames, the change in light level effectively blinds the camera. Depending on how fast the car is going and the outside light level, the blinding effect can be of different magnitude.\\ 
    To simulate this we create a 3x3 Gaussian kernel:
    \begin{equation}
    K = \dfrac{1}{16}   
        \begin{bmatrix}
        1 & 2 & 1\\
        2 & 4 & 2\\
        1 & 2 & 1
        \end{bmatrix}
    \end{equation}
    To create a high exposure effect, where too many light rays hit the sensor, we multiply the kernel by a factor of $(1+3 \times N_{lvl})$. To create the opposite effect, we divide the kernel by the same value. When convolving the kernel with the original image, this results in higher or lower pixel intensity, with a smoothing effect that adds realism to the generated image.
    Figure \ref{hi-lo_exp} shows these effects at different distortion levels.
    
    \begin{figure}[]
    \centerline{\includegraphics[width=0.5\textwidth]{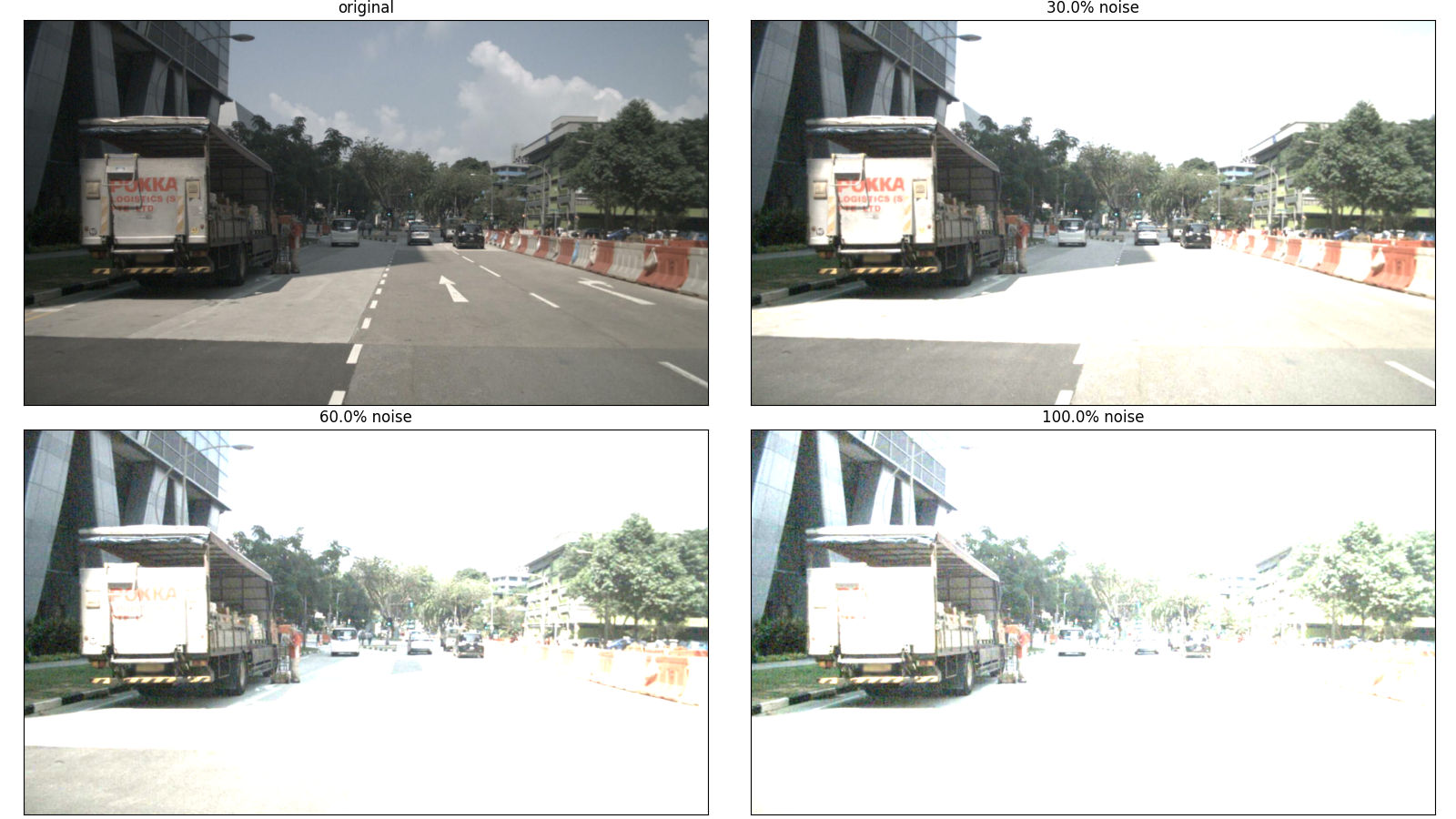}}
    \centerline{\includegraphics[width=0.5\textwidth]{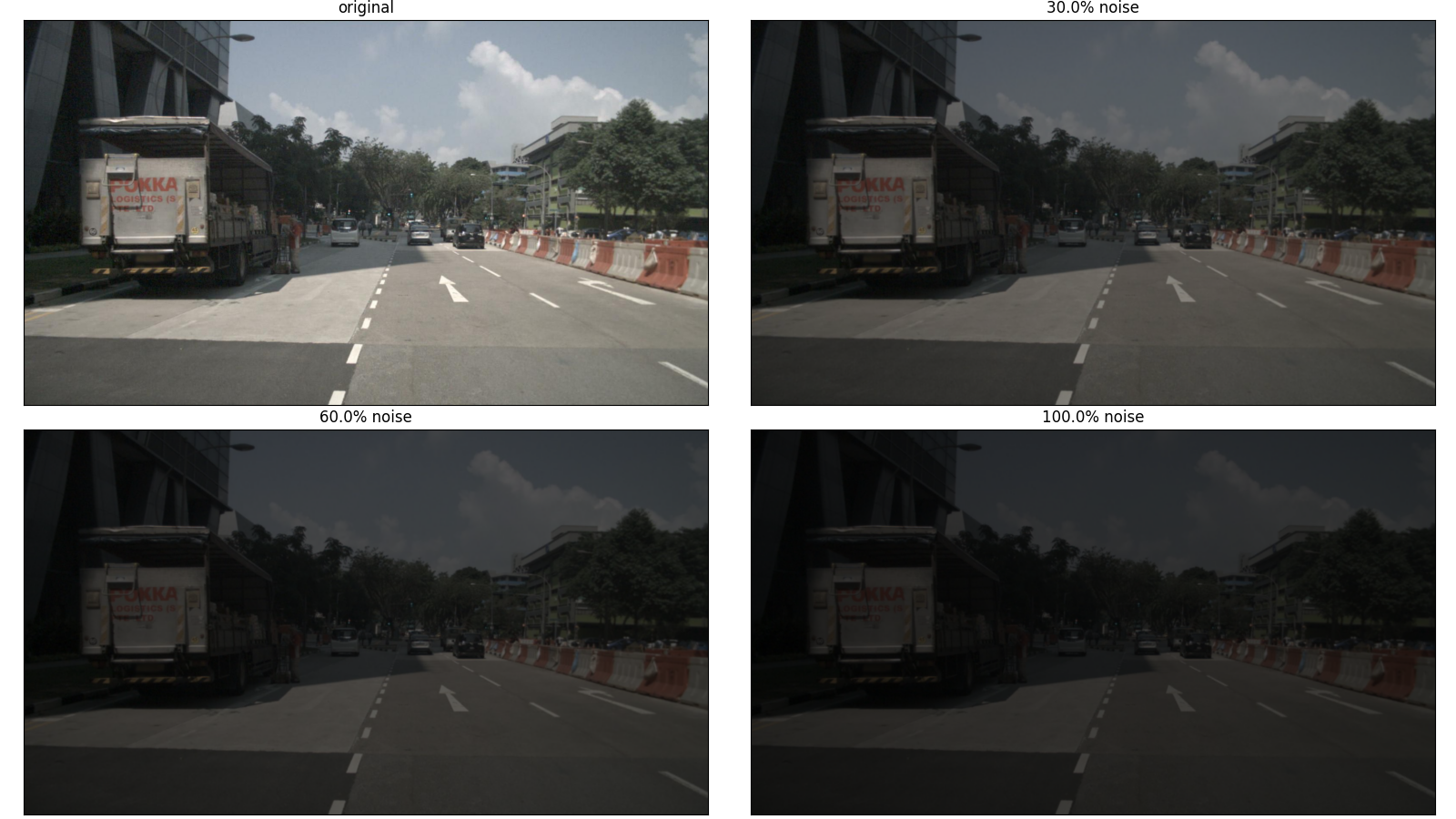}}
    \caption{High (top) and Low (bottom) exposure effect at 0\%, 30\%, 60\%, and 100\% (max) distortion levels.}
    \label{hi-lo_exp}
    \end{figure}
   
    \item \textbf{Additive Noise.} Perhaps the most commonly experienced degradation, additive noise creates a grain-like effect on images. This may happen due to EM noise, thermal noise inside the sensor, or even in low-intensity light level scenarios such as during nighttime, when the quality of the image gets strongly degraded. While this can be filtered out by a Gaussian Kernel, one must be careful as the more an image gets filtered the burrier it becomes. Filtering the right amount of times or with the right kernel would also benefit from knowledge of the noise level.\\
    This is commonly simulated by drawing $W \times H$ values from a Normally distributed random variable $N(0,\sigma)$, with $W$ and $H$ being the width and height of the image, respectively, and adding the resulting noise map to the original image's pixels. We directly set $\sigma=N_{lvl}$ to control the amount of noise. Figure \ref{add_noise} shows the additive Gaussian noise effect at different distortion levels.
    
    \begin{figure}[]
    \centerline{\includegraphics[width=0.5\textwidth]{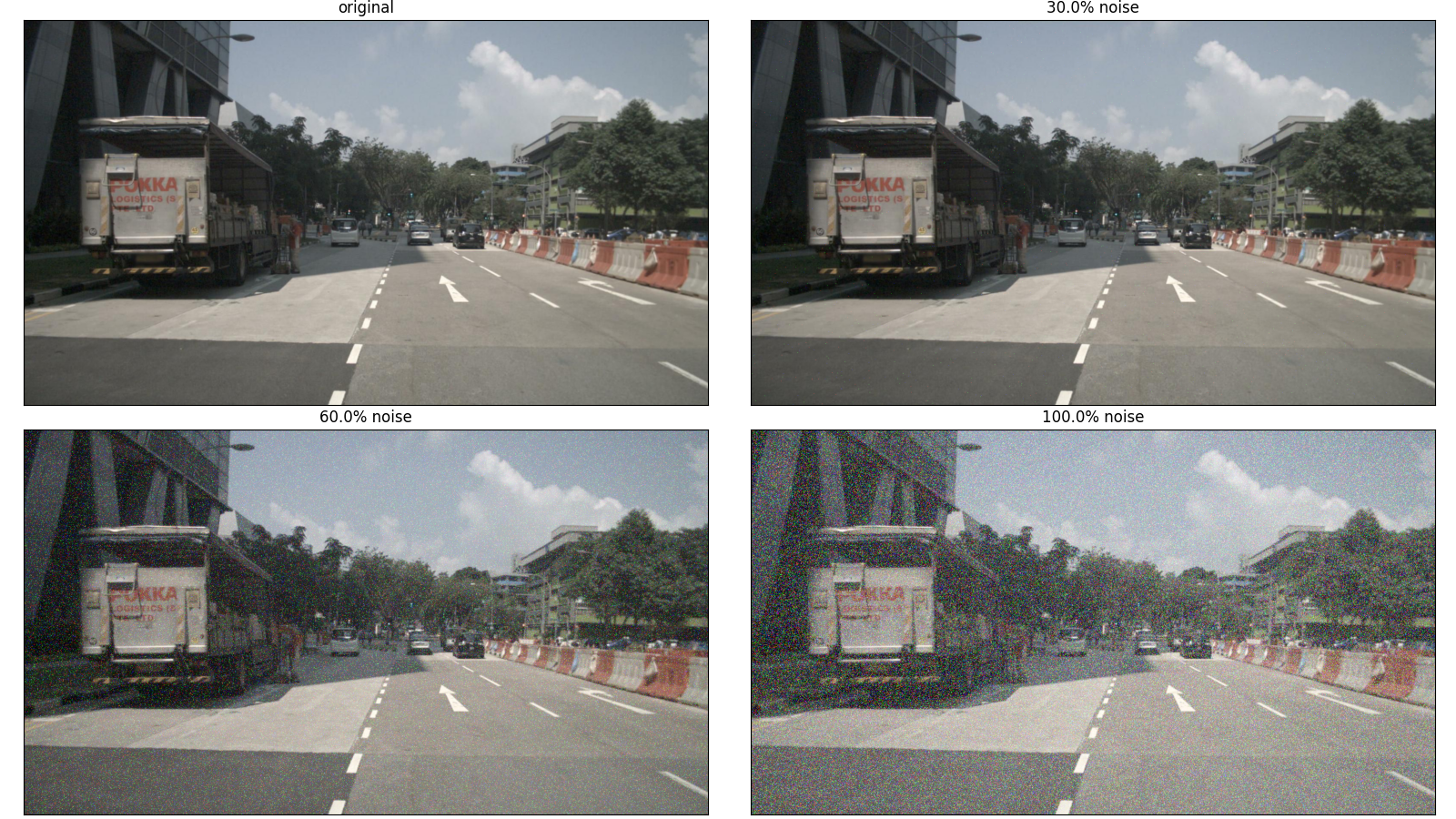}}
    \caption{Normally distributed additive noise effect at 0\%, 30\%, 60\%, and 100\% (max) distortion levels}
    \label{add_noise}
    \end{figure}
\end{enumerate}

\subsection{Synthetic Radar Degradation}
For the radar data, we must first focus on the sensor used by nuScenes: Continental's ARS 408-21 Long Range Radar Sensor 77GHz. The output data given by the sensor at each sweep is a 3D point-cloud with coordinates $(x,y,z)$, a relative velocity vector $(v_x,v_y)$, a motion-compensated relative radial velocity vector $(V_{x_{comp}},V_{y_{comp}})$, a RCS (Radar Cross Section) value, and miscellaneous fields giving information on the false alarm probability of a detection, its estimated dynamic properties, cluster validity state, and the state of its Doppler ambiguity solution. \\
Here we can note a few items: 
\begin{itemize}
    \item The motion compensated velocity is a \textit{radial} velocity, i.e. the projection of the actual motion compensated velocity along the line between the sensor's center and the point. On a similar note, the velocity $(v_x,v_y)$ is \textit{not} a radial velocity, and thus is being estimated by the radar, probably using tracking methods, as radars can only measure a radial velocity, using the Doppler effect. Unfortunately we cannot know exactly how this estimation is being done as each manufacturer has its own method.
    \item Similarly, most of the miscellaneous fields are estimated internally using unknown methods.
    \item Each point extracted by the radar is actually a cluster of points, the values are thus averaged out. This is most probably done to remove outliers and maximize confidence rates.
    \item Finally, in nuScenes the radar is set with a 0° elevation angle, so the points position is always $(x,y,0.0)$.\\
\end{itemize}

\textbf{Noise level:}\\
In signal processing terminology, noise is often referred as a decrease in dB of the overall Signal to Noise Ratio (SNR) defined by:\\
\begin{equation}
    SNR_{lin} = \dfrac{P_r}{P_N}
\end{equation}
 With $P_r$ the received signal's power, and $P_N$ the noise power. We denote $SNR_{lin}$ the SNR in linear scale and $SNR_{dB}$ the SNR in dB scale.
 For the radar data, the noise level $N_{lvl}$ is defined so that a noise level of N\% is a decrease by $N/10$ dB. Thus:
 \begin{equation}
    SNR'_{dB} = SNR_{dB} - \dfrac{N}{10} dB 
\end{equation}
With $SNR'$ being the new Signal to Noise Ratio after synthesis.
In this way, a 100\% noise level implies a -10 dB decrease, and a 0\% noise level implies a 0 dB decrease of SNR. In linear scale this translates to:\\
\begin{equation}
    \begin{split}
    SNR'_{lin} &= 10^{SNR'_{dB}/10} \\
    &= 10^{SNR_{dB}/10} . 10^{- N/100} \\
    &= SNR_{lin} .  10^{- \dfrac{N}{100}}
    \end{split}
\end{equation}
In other terms:\\
\begin{equation}
 \label{snr_eq}
 \dfrac{SNR'_{lin}}{SNR_{lin}} = 10^{- \dfrac{N_{lvl}}{100}}
\end{equation}

To realistically synthesize degraded radar data, we divide our algorithm in 3 steps: Ghost points generation, False negatives, and noise-induced range/Doppler shifts.\\

\textbf{a) Ghost points generation.} In some urban environments, multipath artifacts resulting in ghost points (or false positives) may appear more frequently than in nuScenes'. In some cases, they may even form a large enough cluster and be picked up by the radar if the resulting point's RCS value is high enough. To simulate a higher apparition rate of ghost points due to highly reflective environments, we randomly draw at each frame a number between 0 and a fixed value (set at 4) to determine how many points will be generated. \\
For each ghost points, we draw a range/angle couple $(r,\theta)$ from two uniform random variables $U_r(r_{min},r_{max})$ and $U_\theta(\theta_{min},\theta_{max})$. We set $r_{min}$ to be the radar's minimum rage of 0.2m, and $r_{max}$ to be the farthest point in the current frame, with an added margin of 10m. The reason behind that is while the maximum range of the radar is 250m, most points lie below 110m from observed point clouds. A single ghost point that stands far away from the other clusters would be very easily identifiable and does not represent a real threat to the integrity of the point cloud for a detection network. $\theta_{min}$ and $\theta_{max}$ depend on the selected range, following the limits of the radar's physical detection ranges, which change between short range (under 10m), mid range (under 100m), and far range (above 100m). As mentioned before, there is no way for us to know how the radar estimates the relative velocity of a point, so we sample one from the current frame. As we now have the relative velocity vector $\vec{V}=(v_x,v_y)$, we can project it on the line of sight vector $\vec{r}$ to create a synthetic radial relative velocity vector $\vec{V_r}$. From there we re-create a synthetic compensated radial velocity vector $\vec{V_{comp}}$ knowing the velocity of the nuScenes vehicle $\vec{V_{ego}}$:\\
\begin{equation}
 \vec{V_{comp}} = (\mathbf{V_r} - \mathbf{V_{ego}}).\vec{V_r}
\end{equation}
with $\mathbf{V_r}$ and $\mathbf{V_{ego}}$ the magnitudes of the radial relative velocity vector and of the ego velocity vector (respectively).\\
Finally, we generate a synthetic RCS value by sampling from the known RCS distribution of the current frame, using a bounded half Gaussian variable $X\sim|N(0,1/3)|$ to generate low values more often while preserving the possibility of getting higher ones. The number drawn from $X$ is then mapped to the ordered distribution of RCS values in the current frame. While ghost points tend to be of low RCS, we could sometimes get a very high-RCS cluster that yields a high-RCS valued multipath aberration. The resulting ghost point would have a high RCS comparatively to other points.\\
We also randomly draw a point-state value, which is the radar-estimate state of a point, between selected possible values: low RCS, high child probability, high probability of being a 50° artifact, high probability of being an artifact, and no local maximum. Effectively we consider there is high probability the sensor would estimate something is wrong with the point, but it may mislabel what it is.\\

\textbf{b) RCS-ruled False Negatives.} As the noise power increases (and SNR decreases), the amount of missed points increases in consequence as they disappear behind the background noise. Since we have access to the RCS value, we can use the radar equation:
\begin{equation}
 P_r = \dfrac{P_t G_t A_{eff}}{(4\pi)^2r^4}\sigma
\end{equation}
With $P_r$ the received power, $P_t$ the transmitted power, $G_t$ the gain of the antenna, $A_{eff}$ the effective aperture of the received antenna (in our case the antenna is transmitter/receiver), $\sigma$ the RCS of the point, and $r$ its range.\\
What this effectively means is at a given frame, since $P_t$, $G_t$ and $A_{eff}$ are constant values, the received power($P_r$) is proportional to the RCS ($\sigma$) divided by the range ($r$) of a point to the power of four:\\
\begin{equation}
 P_r \propto \dfrac{\sigma}{r^4}
\end{equation}
Which implies:
\begin{equation}
 SNR \propto \dfrac{\sigma}{r^4}
\end{equation}
And thus from \ref{snr_eq}:
\begin{equation}
 SNR' \propto \dfrac{\sigma}{r^4} \times 10^{-N/100}
\end{equation}
Based on the reasonable assumption that the lowest RCS value of a given frame (represented by $\beta_{min}$) is the minimum detection threshold set by the sensor, we can calculate a coefficient $\alpha = \dfrac{\sigma}{r^4} \times 10^{-N/100} + w$, with $w \sim N(0,\beta_{min})$ a small random noise representing physical uncertainty and RCS estimation error. Any $\alpha$ under $\beta_{min} = min(\dfrac{\sigma}{r^4})$ is removed from the point cloud. The higher $N_{lvl}$ gets, the more points are dropped from the dataset. From observations, less than 10 points are removed with a small $N_{lvl}$ and more than 60 can be cut out of the dataset at -10dB noise level.

\begin{figure}[htbp]
    \centerline{\includegraphics[width=0.5\textwidth]{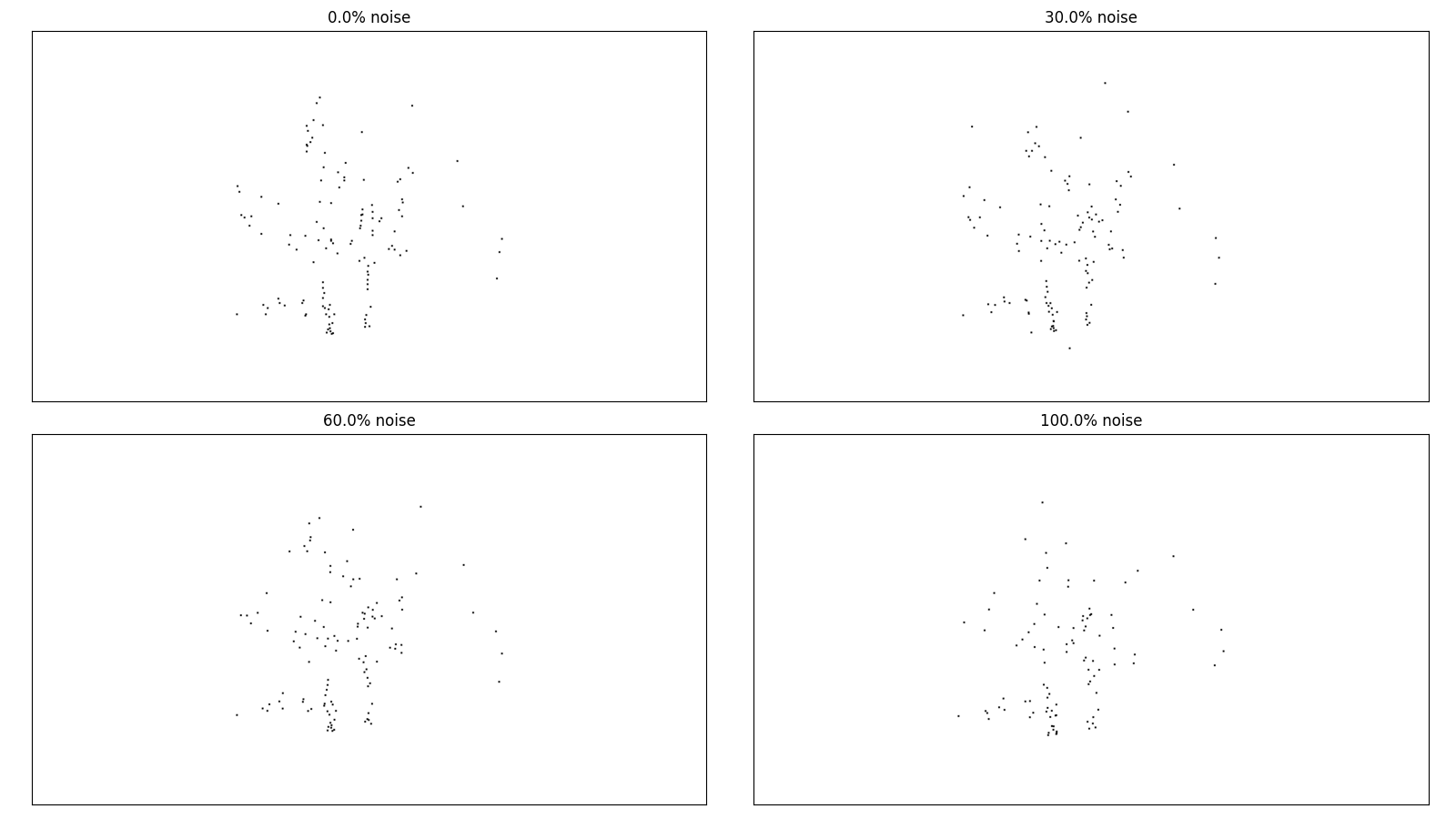}}
    \caption {Effects of noise synthesizer on the radar point cloud in a bird-eye-view (top-down) projection at 0dB, -3dB, -6dB, and -10dB Signal to Noise Ratio.}
    \label{figradaro3d}
\end{figure}

\begin{figure}[htbp]
    \centerline{\includegraphics[width=0.5\textwidth]{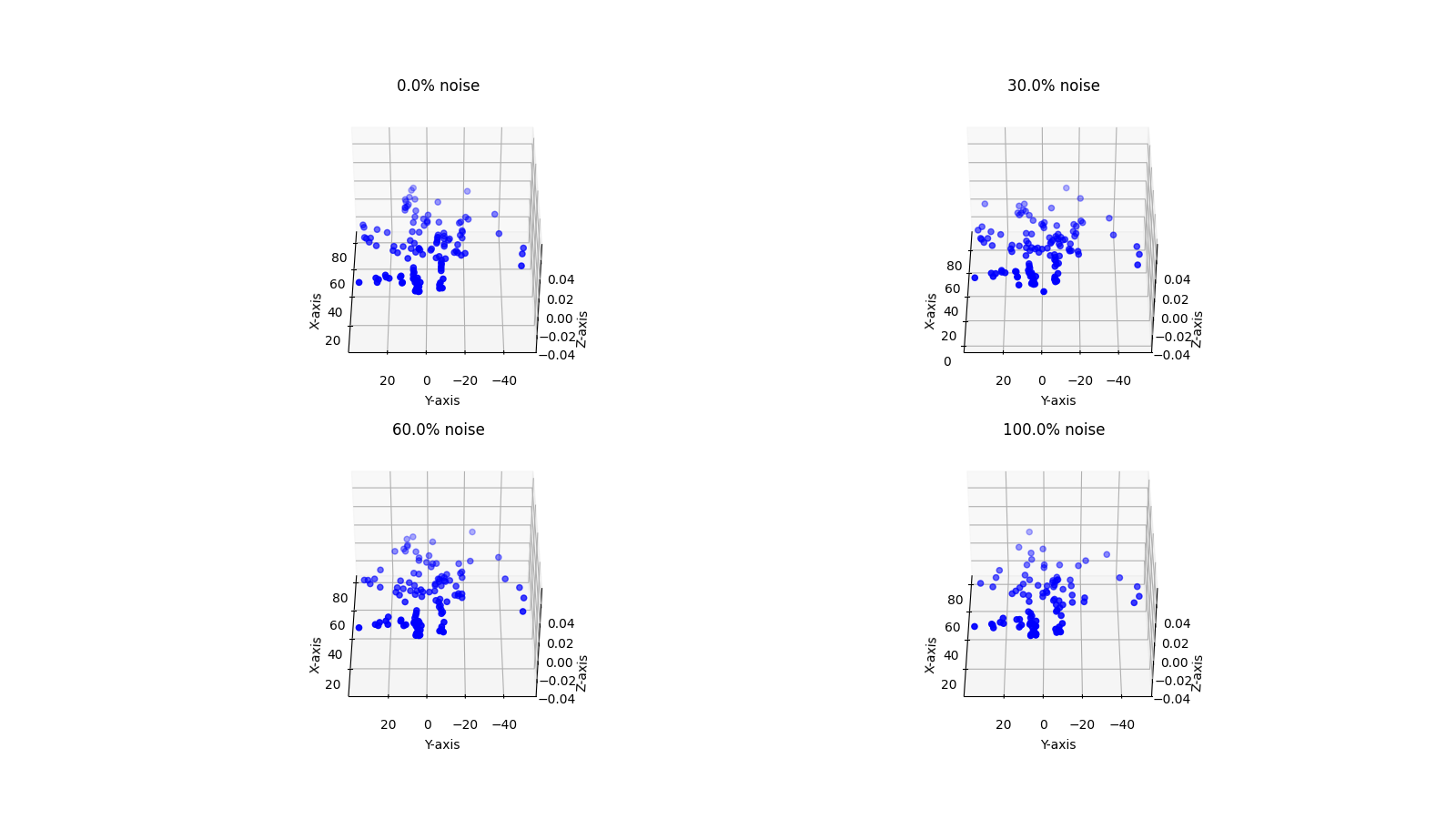}}
    \caption{Effects of noise synthesizer on the radar point cloud in a 3D scan at 0dB, -3dB, -6dB, and -10dB Signal to Noise Ratio.}
    \label{figradarplt}
\end{figure}

\textbf{c) Noise-induced shifts.} As noise level increases (and SRN decreases), the measuring accuracy of a sensor increases as well. In fact, the Cramér-Rao bound suggests \cite{principlesofmodernradar681}:
\begin{equation}
\label{cramerrao}
 acc \propto \dfrac{1}{\sqrt{SNR}} 
\end{equation}
Furthermore, the accuracy values for range, angle, and velocity measurements provided by a manufacturer are usually for high-RCS valued targets, in a controlled environment. The accuracy of the actual measurement for each point is going to be different as the SNR value changes. We consider that $SNR \propto RCS$, and recalculate the accuracy for each measurement of each point using \ref{cramerrao}.\\
From \ref{snr_eq}, we get:\\
\begin{equation}
 acc' \propto \dfrac{1}{\sqrt{SNR'}} = \dfrac{1}{\sqrt{10^{-N/100}}}  
\end{equation}
Finally, we draw noise values $w_r,w_\theta,w_v$ from normally distributed random variables $N(0,acc'_r),N(0,acc'_\theta),N(0,acc'_v)$ and add them to the original $r,\theta,V$ values. We also propagate the noise on the compensated radial relative velocity $V_{comp}$.
Figure \ref{figradaro3d} shows the effect of different noise levels on the radar point cloud in a bird-eye-view (top-down) projection, and Figure \ref{figradarplt} shows the same point clous in a 3D scan. We see less points at lower SNR, some ghost point appearing in close range at -3 dB, and overall bigger position shifts at lower SNR.

\section{Results on Noise Level Detection}
We create a simple neural network to analyze the synthetic dataset. The camera analyzer is a U-NET \cite{unet} of 7 layers (4 in, 3 out), leveraging the encoder/decoder approach of these architecture to efficiently extract noise features. The radar analyzer is made of three 1D convolutional layers followed by two dense layers. Both network are activated by a single dense layer to produce 11 outputs, corresponding to noise levels going from 0 to 100\% with a step size of 10\%.\\
For the cameras our training set is composed of 2 scenes sampled randomly from nuScenes' mini dataset, and our validation and test data are each composed of one other scene. Note here that we do not use the night data in our camera training, as it is considered as an unlabeled adverse scenario. In both camera and radar training, we also only used ``keyframes" (taken at 2Hz) as annotations of the others are interpolated.\\

After applying the different types of noise at each level on the images, the resulting training set size is of 19680 images, plus 10086 for validation, and 10086 for testing.\\

For the radar point clouds, we do the same process, but with 3 training scenes instead of 2. The resulting training set size is of 6501 clouds, while validation and testing are 2255 and 2145 respectively.\\

Training was done on a single NVIDIA GeForce RTX 3060.\\

\begin{figure}[htbp]
    \centerline{\includegraphics[width=0.5\textwidth]{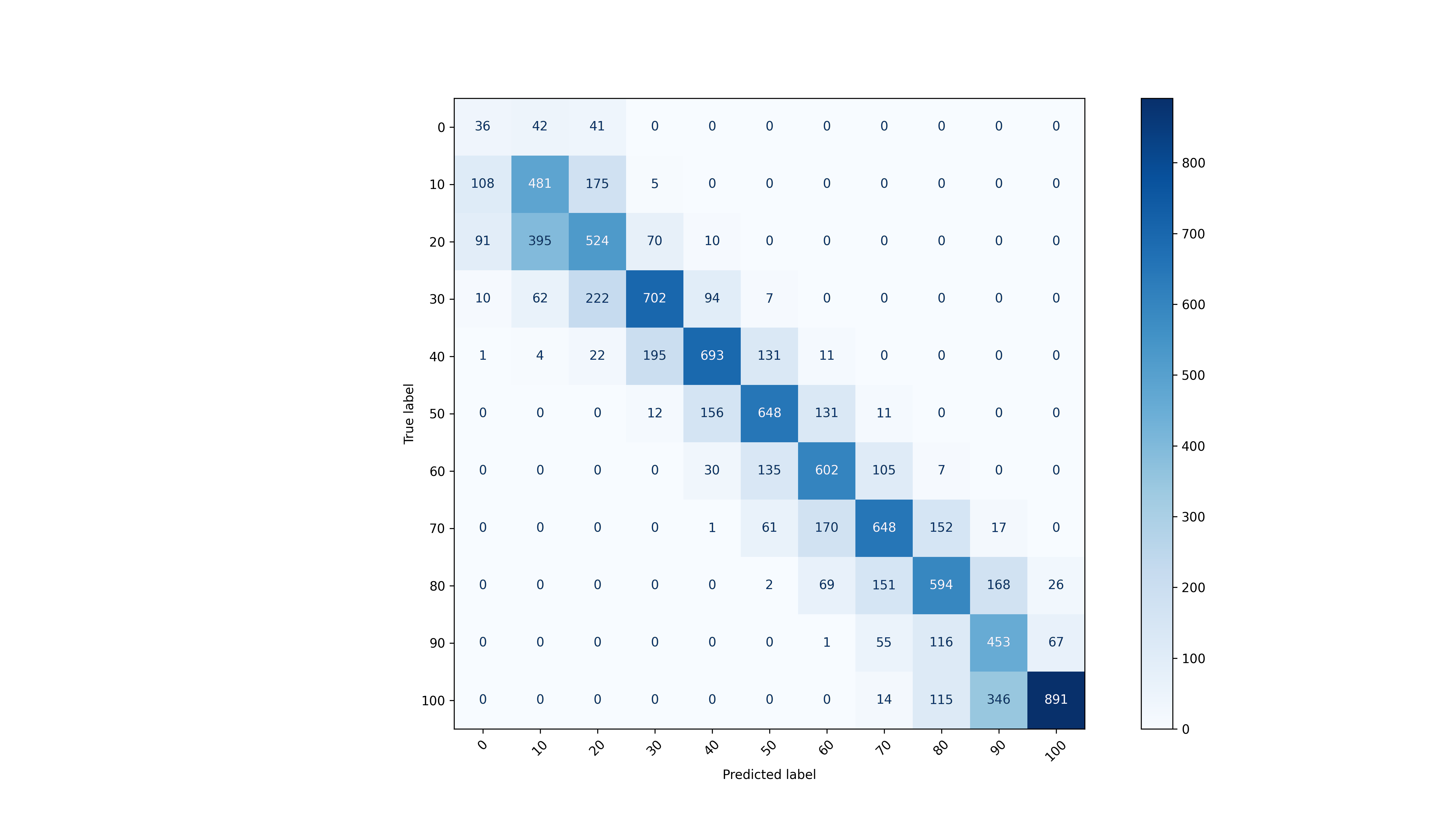}}
    \caption{Confusion matrix for our Image Noise Recognition Network}
    \label{confmat_conv}
\end{figure}

\begin{figure}[htbp]
    \centerline{\includegraphics[width=0.5\textwidth]{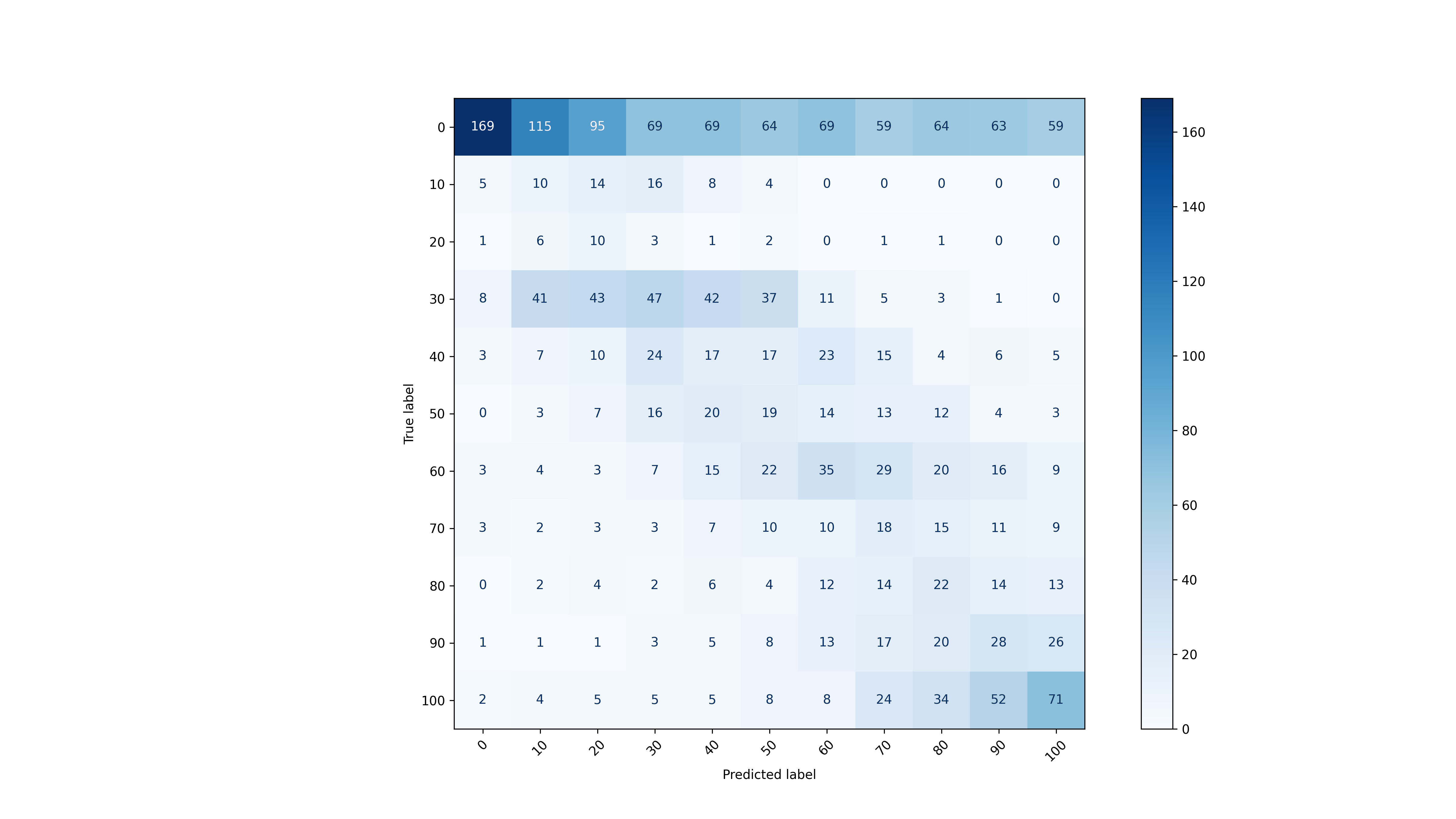}}
    \caption{Confusion matrix for our Radar Noise Recognition Network}
    \label{confmat_radar}
\end{figure}

Our results are in Table \ref{results}, with the networks' confusions matrices in Figure \ref{confmat_conv} and Figure \ref{confmat_radar}.
\begin{table}[!h]
\centering
\renewcommand{\arraystretch}{1.0}
\small
\begin{tabular}{c|c|c|c|c}
\toprule
Data type   & Accuracy (\%) & TP $\uparrow$ & FP $\uparrow$     & Labels\\
\midrule
Camera      & 62.19         & 6272          & 3814              & 10086 \\
Radar       & 20.79         & 446          & 1699              & 2145 \\
\midrule
Total       & 54.4         & 6718          & 5513              & 12341 \\
\bottomrule
\end{tabular}
\caption{Test data results for our noise recognition network on camera and radar data.}
\label{results}
\end{table}

\section{Conclusion and Further works}
In conclusion this papers presents a realistic data synthesizer for autonomous vehicle camera and radar datasets, focusing on simulating sensor failures and data deterioration due to internal or external noise. We also show we can train a light Noise Recognition model able to quantify the degradation of data in both sensors. Our approach allows for plug-and-play use on any other existing method with the possibility to switch to another model at certain noise levels, or try to automatically filter out the noise with this knowledge.
Future works include adding adverse conditions such as weather simulation for the camera data or jammer for the radar data. Furthermore, our noise recognition model is a baseline model and future work include making it lighter, faster, and more accurate. In this regard, we currently limited our training to 5 scenes out of the 1000 provided by nuScenes.

\section*{Acknowledgment}

This project could not have been realized without Dr. Nathan Goodman, professor at the ECE school of the university of Oklahoma, and his help on dealing with the radar data synthesis. 

%%%%%%%%% REFERENCES 
{\small
\bibliographystyle{IEEEtran}
\bibliography{main}
}

\end{document}